\documentclass[11pt,a4paper]{article}
\usepackage[hyperref]{acl2018}
\usepackage{times}
\usepackage{latexsym}
\usepackage{url}
\usepackage{amssymb}
\usepackage{amsmath}
\usepackage{xcolor}
\usepackage{booktabs}
\usepackage{tabulary}
\usepackage{color}
\usepackage{graphicx}
\usepackage{float}

\aclfinalcopy

\newcommand{\lang}[1]{\textsc{#1}}
\newcommand{\cat}{\lang{cat}}
\newcommand{\ces}{\lang{ces}}
\newcommand{\deu}{\lang{deu}}
\newcommand{\eng}{\lang{eng}}
\newcommand{\jpn}{\lang{jpn}}
\newcommand{\spa}{\lang{spa}}
\newcommand{\zho}{\lang{zho}}

\title{Polyglot Semantic Role Labeling}
  
\author{Phoebe Mulcaire$^{\heartsuit}$ ~~ Swabha
  Swayamdipta$^{\diamondsuit}$  ~~ \textbf{Noah A. Smith}$^{\heartsuit}$
\\
$^{\heartsuit}$Paul G. Allen School of Computer Science \& Engineering, 
University of Washington \\
$^\diamondsuit$School of Computer Science, Carnegie Mellon University\\
{\tt \{pmulc,nasmith\}@cs.washington.edu, swabha@cs.cmu.edu} 
}

\date{}

\begin{document}
\maketitle

\begin{abstract}
Previous approaches to multilingual semantic dependency parsing treat languages independently, without exploiting the similarities between semantic structures across languages.
We experiment with a new approach where we combine resources from a pair of languages in the CoNLL 2009 shared task \cite{hajivc2009conll} to build a  \textit{polyglot} semantic role labeler.
Notwithstanding the absence of parallel data, and the dissimilarity in annotations between languages, our approach results in an improvement in SRL performance on multiple languages over a monolingual baseline.
Analysis of the polyglot model shows it to be advantageous in  lower-resource settings.
\end{abstract}

\section{Introduction}
The standard approach to multilingual NLP is to design a single architecture, but tune and train a separate model for each language. 
While this method allows for customizing the model to the particulars of each language and the available data, it also presents a problem when little data is available: extensive language-specific annotation is required. The reality is that most languages have very little annotated data for most NLP tasks.

\newcite{ammar2016malopa} found that using training data from multiple languages annotated with Universal Dependencies \cite{nivre2016universal}, and represented using multilingual word vectors, outperformed monolingual training. 
Inspired by this, we apply the idea of training one model on multiple languages---which we call polyglot training---to PropBank-style semantic role labeling (SRL).
We train several parsers for each language in the CoNLL 2009 dataset \cite{hajivc2009conll}:  a traditional monolingual version, and variants which additionally incorporate supervision from English portion of the dataset.
To our knowledge, this is the first multilingual SRL approach to combine supervision from several languages.

The CoNLL 2009 dataset includes seven different languages, allowing study of trends across the same.
Unlike the Universal Dependencies dataset, however, the semantic label spaces are entirely language-specific, making our task more challenging.
Nonetheless, the success of polyglot training in this setting demonstrates that sharing of statistical strength across languages does not depend on explicit alignment in annotation conventions, and can be done simply through parameter sharing.
We show that polyglot training can result in better labeling accuracy than a monolingual parser, \emph{especially} for low-resource languages.
We find that even a simple combination of data is as effective as more complex kinds of polyglot training.
We include a breakdown into label categories of the differences between the monolingual and polyglot models.
Our findings indicate that polyglot training consistently improves label accuracy for common labels.

\section{Data}
\label{sect:data}

\begin{figure}
\vspace{0.1cm}
\includegraphics[scale=0.225]{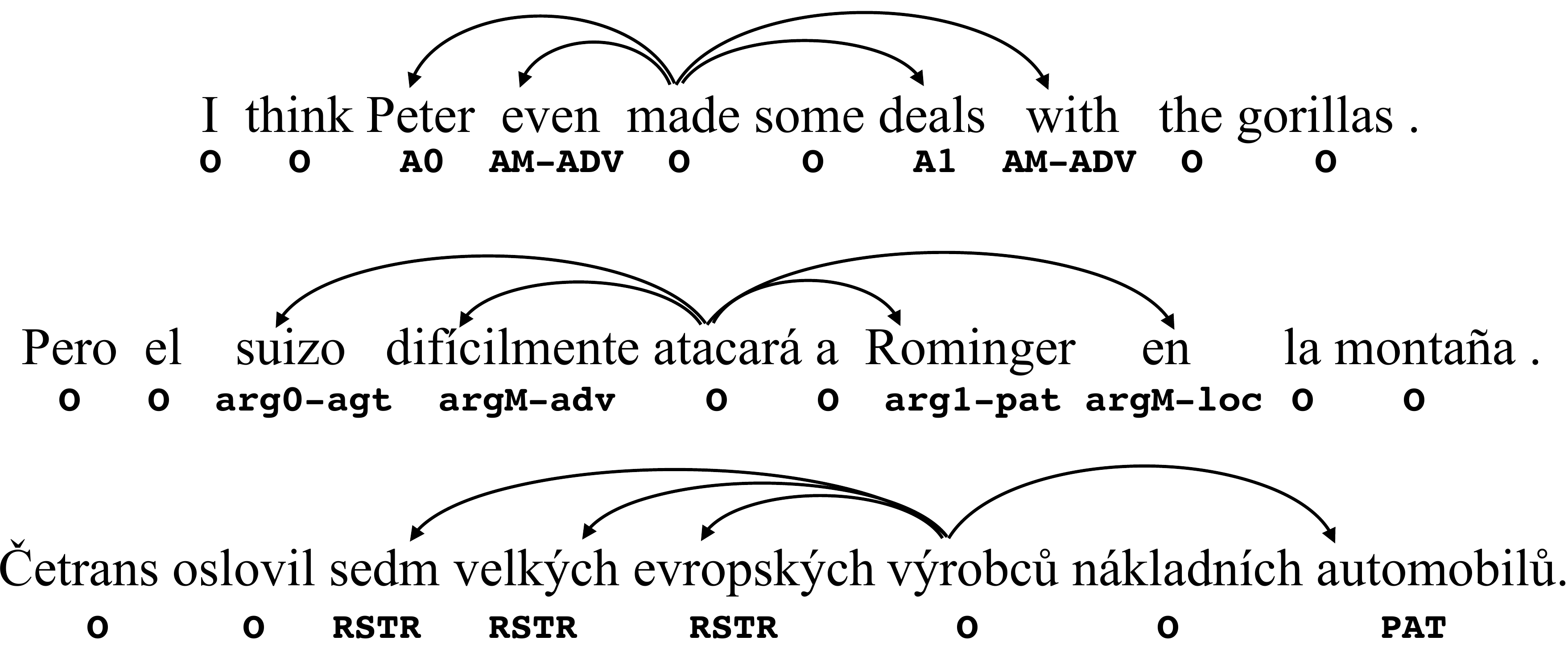}
\caption{Example predicate-argument structures from English, Spanish, and Czech. Note that the argument labels are different in each language.}
\vspace{-.5cm}
\label{fig:data}
\end{figure}

We evaluate our system on the semantic role labeling portion of the CoNLL-2009 shared task \cite{hajivc2009conll}, on all seven languages, namely Catalan, Chinese, Czech, English, German, Japanese and Spanish.
For each language, certain tokens in each sentence in the dataset are marked as predicates.
Each predicate takes as arguments other words in the same sentence, their relationship marked by labeled dependency arcs.
Sentences may contain no predicates.

Despite the consistency of this format, there are significant differences between the training sets across languages.\footnote{This is expected, as the datasets were annotated independently under diverse formalisms and only later converted into CoNLL format \cite{hajivc2009conll}.}
English uses PropBank role labels~\cite{Palmer:05}. Catalan, Chinese, English, German, and Spanish include (but are not limited to) labels such as ``arg$_0$-agt'' (for ``agent'') or ``A$_0$'' that may correspond to some degree to each other and to the English roles. 
Catalan and Spanish share most labels (being drawn from the same source corpus, AnCora; \citealp{taule2008ancora}), and English and German share some labels.
Czech and Japanese each have their own distinct sets of argument labels, most of which do not have clear correspondences to English or to each other.

We also note that, due to semi-automatic projection of annotations to construct the German dataset, more than half of German sentences do \emph{not} include labeled predicate and arguments. 
Thus while German has almost as many sentences as Czech, it has by far the fewest training examples (predicate-argument structures); see Table \ref{tab:datasets}.

\begin{table}[t!]
\center
\begin{tabulary}{\columnwidth}{@{}lrrr@{}}
\toprule
& \# sentences	& \shortstack{\# sentences w/\\1+ predicates} & \# predicates 	\\ 
\midrule
\midrule[.03em]
\cat					&13200 			& 12876	& 37444		\\
\ces					& 38727			& 38579	& 414133		 \\
\deu					& 36020			& 14282 & 17400			\\
\eng					& 39279			& 37847	& 179014			\\
\jpn					& 4393  		& 4344  & 25712			\\
\spa					& 14329			& 13836	& 43828			\\
\zho					& 22277			& 21073	& 102827			\\
\bottomrule
\bottomrule
\end{tabulary}
\caption{Train data statistics. Languages are indicated with ISO 639-3 codes.}
\label{tab:datasets}
\vspace{-.4cm}
\end{table}

\section{Model}
\label{sect:model}

Given a sentence with a marked predicate, the CoNLL 2009 shared task requires disambiguation of the sense of the predicate, and labeling all its dependent arguments.
The shared task assumed predicates have already been identified, hence we do not handle the predicate identification task.

Our basic model adapts the span-based dependency SRL model of \newcite{He2017-deep_srl}.
This adaptation treats the dependent arguments as argument spans of length 1.
Additionally, BIO consistency constraints are removed from the original model--- each token is tagged simply with the argument label or an empty tag. 
A similar approach has also been proposed by \newcite{marcheggiani2017lstm}.

The input to the model consists of a sequence of pretrained embeddings for the surface forms of the sentence tokens.
Each token embedding is also concatenated with a vector indicating whether the word is a predicate or not.
Since the part-of-speech tags in the CoNLL 2009 dataset are based on a different tagset for each language, we do not use these.
Each training instance consists of the annotations for a single predicate.
These representations are then passed through a deep, multi-layer bidirectional LSTM~\cite{Graves:13,Hochreiter:97} with highway connections \cite{srivastava2015training}.

We use the hidden representations produced by the deep biLSTM for both argument labeling and predicate sense disambiguation in a multitask setup; this is a modification to the models of \newcite{He2017-deep_srl}, who did not handle predicate senses, and of \newcite{marcheggiani2017lstm}, who used a separate model.
These two predictions are made independently, with separate softmaxes over different last-layer parameters; we then combine the losses for each task when training.
For predicate sense disambiguation, since the predicate has been identified, 
we choose from a small set of valid predicate senses as the tag for that token.
This set of possible senses is selected based on the training data:  we map from lemmatized tokens to predicates and from predicates to the set of all senses of that predicate. 
Most predicates are only observed to have one or two corresponding senses, making the set of available senses at test time quite small (less than five senses/predicate on average across all languages). If a particular lemma was not observed in training, we heuristically predict it as the first sense of that predicate. For Czech and Japanese, the predicate sense annotation is simply the lemmatized token of the predicate, giving a one-to-one predicate-``sense'' mapping.

For argument labeling, every token in the sentence is assigned one of the argument labels, or $\textsc{null}$ if the model predicts it is not an argument to the indicated predicate.

\subsection{Monolingual Baseline}
\label{sect:baseline}

We use pretrained word embeddings as input to the model. For each of the shared task languages, we produced GloVe vectors \cite{pennington2014glove} from the news, web, and Wikipedia text of the Leipzig Corpora Collection \cite{goldhahn2012building}.\footnote{For English we used the vectors provided on the GloVe website \url{nlp.stanford.edu/projects/glove/}.} We trained 300-dimensional vectors, then reduced them to 100 dimensions with principal component analysis for efficiency.

\subsection{Simple Polyglot Sharing}
\label{sect:combo}

In the first polyglot variant, we consider multilingual sharing between each language and English by using pretrained \emph{multilingual} embeddings.
This polyglot model is trained on the union of annotations in the two languages.
We use stratified sampling to give the two datasets equal effective weight in training,  and we ensure that every training instance is seen at least once per epoch.

\paragraph{Pretrained multilingual embeddings.}
The basis of our polyglot training is the use of pretrained multilingual word vectors, which allow representing entirely distinct vocabularies (such as the tokens of different languages) in a shared representation space, allowing crosslingual learning \cite{klementiev2012inducing}. 
We produced multilingual embeddings from the monolingual embeddings using the method of \newcite{ammar2016massively}: for each non-English language, a small crosslingual dictionary and canonical correlation analysis was used to find a transformation of the non-English vectors into the English vector space \cite{faruqui2014improving}.

Unlike multilingual word representations, argument label sets are disjoint between language pairs, and correspondences are not clearly defined.
Hence, we use separate label representations for each language's labels.
Similarly, while (for example) \eng:look and \spa:mira may be semantically connected, the senses \texttt{look.01} and \texttt{mira.01} may not correspond.
Hence, predicate sense representations are also language-specific.

\subsection{Language Identification}
\label{sect:langid}
In the second variant, we concatenate a language ID vector to each multilingual word embedding and predicate indicator feature in the input representation. This vector is randomly initialized and updated in training.
These additional parameters provide a small degree of language-specificity in the model, while still sharing most parameters. 

\subsection{Language-Specific LSTMs}
\label{sect:frela}
This third variant takes inspiration from the ``frustratingly easy'' architecture of \newcite{daumeiii2007easy} for domain adaptation. In addition to processing every example with a shared biLSTM as in previous models, we add language-specific biLSTMs that are trained only on the examples belonging to one language. Each of these language-specific biLSTMs is two layers deep, and is combined with the shared biSLTM in the input to the third layer. This adds a greater degree of language-specific processing while still sharing representations across languages.
It also uses the language identification vector and multilingual word vectors in the input.

\section{Experiments}
\label{sect:experiments}

\begin{figure}
\vspace{0.1cm}
\includegraphics[scale=0.34]{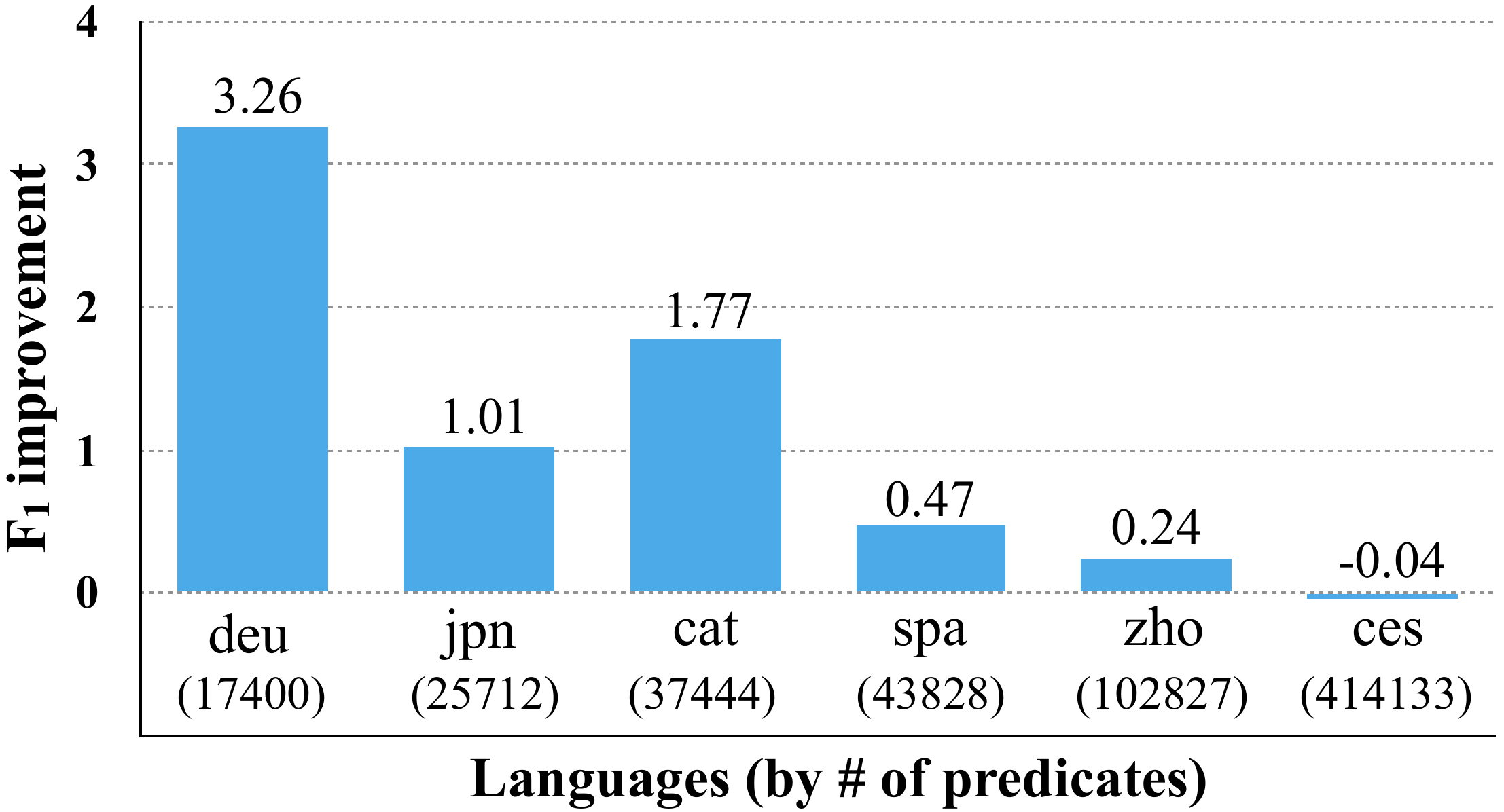}
\vspace{-0.8cm}
\caption{Improvement in absolute $F_1$ with polyglot training with addition of English. Languages are sorted in order of increasing number of predicates in the training set.}
\vspace{-0.4cm}
\label{fig:improve}
\end{figure}

We present our results in Table \ref{tab:comparison}.
We observe that simple polyglot training improves over
monolingual training, with the exception of Czech, where we observe no change in performance. 
The languages with the fewest training examples (German, Japanese, Catalan) show the most improvement, while large-dataset languages such as Czech or Chinese see little or no improvement (Figure \ref{fig:improve}). 

The language ID model performs inconsistently; it is better than the
simple polyglot model in some cases, including Czech, but not in
all. The language-specific LSTMs model performs best on a few
languages, such as Catalan and Chinese, but worst on others. While
these results may reflect differences between languages in the optimal
amount of crosslingual sharing, we focus on the simple polyglot
results in our analysis, which sufficiently demonstrate that
polyglot training can improve performance over monolingual
training.

We also report performance of state-of-the-art systems in each of these languages, all of which make explicit use of syntactic features, \newcite{marcheggiani2017lstm} excepted. 
While this results in better performance on many languages, our model has the advantage of not relying on a syntactic parser, and is hence more applicable to languages with lower resources. 
However, the results suggest that syntactic information is critical for strong performance on German, which has the fewest predicates and thus the least semantic annotation for a semantics-only model to learn from. 
Nevertheless, our baseline is on par with the best published scores for Chinese, and it shows strong performance on most languages.

\begin{table*}[t!]
\centering
\begin{tabulary}{\textwidth}{l|rrrrrrr}
\toprule
\textbf{Model} 			& \cat	& \ces 	& \deu 	& \eng 	& \jpn 	& \spa	& \zho	\\ 
\midrule
\newcite{marcheggiani2017lstm}	&  	-	& 86.00	&  	-	& 87.60	&  	-	& 80.30	& 81.20	\\
Best previously reported		& \textit{80.32} & 86.00	& \textit{80.10}	& \textit{89.10} & \textit{78.15}	& \textit{80.50} & \textit{81.20} \\
\midrule
Monolingual 				& 77.31	& 84.87	& 66.71		& 86.54	& 74.99	& 75.98	& 81.26	\\
+ \eng (simple polyglot)	& 79.08	& 84.82	& 69.97		& --		& 76.00 & 76.45	& 81.50	\\
+ \eng (language ID)		& 79.05	& 85.14 	& 69.49 		& -- 	& 75.77 & 77.32	& 81.42	\\
+ \eng (language-specific LSTMs) & 79.45	& 84.78	& 68.30	& --	& 75.88	& 76.86	& 81.89	\\
\bottomrule
\end{tabulary}
\caption{
	Semantic $F_1$ scores (including predicate sense disambiguation) on the CoNLL 2009 dataset. 
	State of the art for Catalan and Japanese is from \newcite{Zhao2009}, for German and Spanish from \newcite{Roth2016-fn}, for English and Chinese from \newcite{marcheggiani2017gcn}. 
Italics indicate use of syntax.}
\label{tab:comparison}
\vspace{-.2cm}
\end{table*}

\begin{table*}[t!]
\centering
\begin{tabulary}{\textwidth}{l|rrrrrrr}
\toprule
						& \bf arg$_0$	& \bf arg$_1$	& \bf arg$_2$	& \bf arg$_3$	& \bf arg$_4$	& \bf arg$_L$	& \bf arg$_M$	\\
\midrule 
Gold label count (\cat)	& 2117	& 4296	& 1713	& 61	& 71	& 49	& 2968	\\ 
Monolingual \cat ~$F_1$	& 82.06	& 79.06	& 68.95	& 28.89	& 42.42	& 39.51	& 60.85 \\
+ \eng ~improvement		& +2.75	& +2.58	& +4.53	& +18.17& +9.81 & +1.35	& +1.10 \\
\midrule
Gold label count (\spa)	& 2438	& 4295	& 1677	& 49	& 82	& 46	& 3237	\\ 
Monolingual \spa ~$F_1$	& 82.44	& 77.93	& 70.24	& 28.89	& 41.15	& 22.50	& 58.89 \\
+ \eng ~improvement		& +0.37	& +0.43	& +1.35	& -3.40	& -3.48	& +4.01	& +1.26 \\
\bottomrule
\end{tabulary}
\caption{\label{tab:label-breakdown} Per-label breakdown of $F_1$ scores for Catalan and Spanish. These numbers reflect labels for each argument; the combination is different from the overall semantic $F_1$, which includes predicate sense disambiguation.}
\vspace{-.4cm}
\end{table*}

\paragraph{Label-wise results.}
Table \ref{tab:label-breakdown} gives the $F_1$ scores for individual label categories in the Catalan and Spanish datasets, as an illustration of the larger trend.
In both languages, we find a small but consistent improvement in the most common label categories (e.g., arg$_1$ and arg$_M$). 
Less common label categories are sensitive to small changes in performance; they have the largest changes in $F_1$ in absolute value, but without a consistent direction. 
This could be attributed to the addition of English data, which improves learning of representations that are useful for the most common labels, but is essentially a random perturbation for the rarer ones. 
This pattern is seen across languages, and consistently results in overall gains from polyglot training.

One exception is in Czech, where polyglot training reduces accuracy on several common argument labels, e.g., PAT and LOC. 
While the effect sizes are small (consistent with other languages), the overall $F_1$ score on Czech decreases slightly in the polyglot condition. 
It may be that the Czech dataset is too large to make use of the comparatively small amount of English data, or that differences in the annotation schemes prevent effective crosslingual transfer.

Future work on language pairs that do not include English could provide further insights. 
Catalan and Spanish, for example, are closely related and use the same argument label set (both being drawn from the AnCora corpus) which
would allow for sharing output representations as well as input tokens and parameters. 

\paragraph{Polyglot English results.}
For each language pair, we also evaluated the simple polyglot model on the English test set from the CoNLL 2009 shared task (Table \ref{tab:eng-results}).
English SRL consistently benefits from polyglot training, with an increase of 0.25--0.7 absolute $F_1$ points, depending on the language.
Surprisingly, Czech provides the smallest improvement, despite the large amount of data added; the absence of crosslingual transfer in both directions for the English-Czech case, breaking the pattern seen in other languages, could therefore be due to differences in annotation rather than questions of dataset size.

\begin{table*}[t!]
	\centering
	\begin{tabulary}{\textwidth}{rrrrrrr}
\toprule
		\eng -only	& +\cat		& +\ces		& +\deu		& +\jpn		& +\spa		& +\zho	\\
\midrule
		86.54		& 86.79		& 87.07		& 87.07		& 87.11		& 87.24		& 87.10	\\
\bottomrule
	\end{tabulary}
	\caption{Semantic $F_1$ scores on the English test set for each language pair.}
	\label{tab:eng-results} 
\end{table*}

\paragraph{Labeled vs.~unlabeled $F_1$.}

Table \ref{tab:unlabeled} provides unlabeled $F_1$ scores for each language pair. As can be seen here, the unlabeled $F_1$ improvements are generally positive but small, indicating that polyglot training can help both in structure prediction and labeling of arguments. The pattern of seeing the largest improvements on the languages with the smallest datasets generally holds here: the largest $F_1$ gains are in German and Catalan, followed by Japanese, with minimal or no improvement elsewhere.

\begin{table*}[t!]
\centering
\begin{tabulary}{\textwidth}{l|rrrrrrr}
\toprule
\textbf{Model} 			& \cat	& \ces 	& \deu 	& \eng 	& \jpn 	& \spa	& \zho	\\ 
\midrule
Monolingual 				& 93.92	& 91.92	& 87.95	& 92.87 	& 85.55	& 93.61	& 87.93	\\
+ \eng  					& 94.09	& 91.97	& 89.01	& --		& 86.17 & 93.65	& 87.90	\\
\bottomrule
\end{tabulary}
\caption{Unlabeled semantic $F_1$ scores on the CoNLL 2009 dataset.}
\label{tab:unlabeled}
\vspace{-.2cm}
\end{table*}

\section{Related Work}
\label{sect:related}
Recent improvements in multilingual SRL can be attributed to neural architectures. 
\newcite{Swayamdipta2016-qt} present a transition-based stack LSTM model that predicts syntax and semantics jointly, as a remedy to the reliance on pipelined models. 
\newcite{Guo2016-zc} and \citet{Roth2016-fn} use deep biLSTM architectures which use syntactic information to guide the composition.
\newcite{marcheggiani2017lstm} use a simple LSTM model over word tokens to tag semantic dependencies, like our model. 
Their model predicts a token's label based on the combination of the token vector and the predicate vector, and saw benefits from using POS tags, both improvements that could be added to our model.
\newcite{marcheggiani2017gcn} apply the recently-developed graph convolutional networks to SRL, obtaining state of the art results on English and Chinese.
All of these approaches are orthogonal to ours, and might benefit from polyglot training.

Other polyglot models have been proposed for semantics. 
\newcite{Richardson2018-ov-naacl} train on multiple (natural language)-(programming language) pairs to improve a model that translates API text into code signature representations. 
\newcite{Duong2017-qy} treat English and German semantic parsing as a multi-task learning problem and saw improvement over monolingual baselines, especially for small datasets. 
Most relevant to our work is \newcite{Johannsen2015-nb}, which trains a polyglot model for \textit{frame}-semantic parsing. In addition to sharing features with multilingual word vectors, they use them to find word translations of target language words for additional lexical features.

\section{Conclusion}
\label{sect:conclusion}
In this work, we have explored a straightforward method for polyglot training in SRL:  use multilingual word vectors and combine
training data across languages.  This allows sharing without crosslingual alignments, shared annotation, or parallel data.
We demonstrate that a polyglot model can outperform a monolingual one for semantic analysis, particularly for languages with less data.

\section*{Acknowledgments}
We thank Luke Zettlemoyer, Luheng He, and the anonymous reviewers for helpful comments and feedback.
This research was supported in part 
by the Defense Advanced Research Projects Agency (DARPA) Information
Innovation Office (I2O) under the Low Resource Languages for Emergent
Incidents (LORELEI) program issued by DARPA/I2O under contract
HR001115C0113 to BBN.  Views expressed are those of the authors alone.

\bibliography{multilingual}

\begin{thebibliography}{24}
\expandafter\ifx\csname natexlab\endcsname\relax\def\natexlab#1{#1}\fi

\bibitem[{Ammar et~al.(2016{\natexlab{a}})Ammar, Mulcaire, Ballesteros, Dyer,
  and Smith}]{ammar2016malopa}
Waleed Ammar, George Mulcaire, Miguel Ballesteros, Chris Dyer, and Noah Smith.
  2016{\natexlab{a}}.
\newblock Many languages, one parser.
\newblock \emph{Transactions of the Association for Computational Linguistics},
  4:431--444.

\bibitem[{Ammar et~al.(2016{\natexlab{b}})Ammar, Mulcaire, Tsvetkov, Lample,
  Dyer, and Smith}]{ammar2016massively}
Waleed Ammar, George Mulcaire, Yulia Tsvetkov, Guillaume Lample, Chris Dyer,
  and Noah~A Smith. 2016{\natexlab{b}}.
\newblock Massively multilingual word embeddings.
\newblock {arXiv:1602.01925}.

\bibitem[{Daume~III(2007)}]{daumeiii2007easy}
Hal Daume~III. 2007.
\newblock Frustratingly easy domain adaptation.
\newblock In \emph{Proceedings of ACL}.

\bibitem[{Duong et~al.(2017)Duong, Afshar, Estival, Pink, Cohen, and
  Johnson}]{Duong2017-qy}
Long Duong, Hadi Afshar, Dominique Estival, Glen Pink, Philip Cohen, and Mark
  Johnson. 2017.
\newblock Multilingual semantic parsing and code-switching.
\newblock In \emph{Proceedings of CoNLL}.

\bibitem[{Faruqui and Dyer(2014)}]{faruqui2014improving}
Manaal Faruqui and Chris Dyer. 2014.
\newblock Improving vector space word representations using multilingual
  correlation.
\newblock In \emph{Proceedings of EACL}.

\bibitem[{Goldhahn et~al.(2012)Goldhahn, Eckart, and
  Quasthoff}]{goldhahn2012building}
Dirk Goldhahn, Thomas Eckart, and Uwe Quasthoff. 2012.
\newblock Building large monolingual dictionaries at the {Leipzig} corpora
  collection: From 100 to 200 languages.
\newblock In \emph{Proceedings of LREC}.

\bibitem[{Graves(2013)}]{Graves:13}
Alex Graves. 2013.
\newblock Generating sequences with recurrent neural networks.
\newblock {arXiv:1308.0850}.

\bibitem[{Guo et~al.(2016)Guo, Che, Wang, Liu, and Xu}]{Guo2016-zc}
Jiang Guo, Wanxiang Che, Haifeng Wang, Ting Liu, and Jun Xu. 2016.
\newblock A unified architecture for semantic role labeling and relation
  classification.
\newblock In \emph{Proceedings of COLING}.

\bibitem[{Haji{\v{c}} et~al.(2009)Haji{\v{c}}, Ciaramita, Johansson, Kawahara,
  Mart{\'\i}, M{\`a}rquez, Meyers, Nivre, Pad{\'o}, {\v{S}}t{\v{e}}p{\'a}nek
  et~al.}]{hajivc2009conll}
Jan Haji{\v{c}}, Massimiliano Ciaramita, Richard Johansson, Daisuke Kawahara,
  Maria~Ant{\`o}nia Mart{\'\i}, Llu{\'\i}s M{\`a}rquez, Adam Meyers, Joakim
  Nivre, Sebastian Pad{\'o}, Jan {\v{S}}t{\v{e}}p{\'a}nek, et~al. 2009.
\newblock The conll-2009 shared task: Syntactic and semantic dependencies in
  multiple languages.
\newblock In \emph{Proceedings of CoNLL}.

\bibitem[{He et~al.(2017)He, Lee, Lewis, and Zettlemoyer}]{He2017-deep_srl}
Luheng He, Kenton Lee, Mike Lewis, and Luke Zettlemoyer. 2017.
\newblock Deep semantic role labeling: What works and what’s next.
\newblock In \emph{Proceedings of ACL}.

\bibitem[{Hochreiter and Schmidhuber(1997)}]{Hochreiter:97}
Sepp Hochreiter and J{\"u}rgen Schmidhuber. 1997.
\newblock Long short-term memory.
\newblock \emph{Neural Computation}, 9(8):1735--1780.

\bibitem[{Johannsen et~al.(2015)Johannsen, Alonso, and
  S{\o}gaard}]{Johannsen2015-nb}
Anders Johannsen, H{\'e}ctor~Mart{\'\i}nez Alonso, and Anders S{\o}gaard. 2015.
\newblock Any-language frame-semantic parsing.
\newblock In \emph{Proceedings of EMNLP}.

\bibitem[{Klementiev et~al.(2012)Klementiev, Titov, and
  Bhattarai}]{klementiev2012inducing}
Alexandre Klementiev, Ivan Titov, and Binod Bhattarai. 2012.
\newblock Inducing crosslingual distributed representations of words.
\newblock \emph{Proceedings of COLING}.

\bibitem[{Marcheggiani et~al.(2017)Marcheggiani, Frolov, and
  Titov}]{marcheggiani2017lstm}
Diego Marcheggiani, Anton Frolov, and Ivan Titov. 2017.
\newblock A simple and accurate syntax-agnostic neural model for
  dependency-based semantic role labeling.
\newblock {arXiv:1701.02593}.

\bibitem[{Marcheggiani and Titov(2017)}]{marcheggiani2017gcn}
Diego Marcheggiani and Ivan Titov. 2017.
\newblock Encoding sentences with graph convolutional networks for semantic
  role labeling.
\newblock In \emph{Proceedings of EMNLP}, Copenhagen, Denmark.

\bibitem[{Nivre et~al.(2016)Nivre, de~Marneffe, Ginter, Goldberg, Hajic,
  Manning, McDonald, Petrov, Pyysalo, Silveira, Tsarfaty, and
  Zeman}]{nivre2016universal}
Joakim Nivre, Marie-Catherine de~Marneffe, Filip Ginter, Yoav Goldberg, Jan
  Hajic, Christopher~D. Manning, Ryan~T. McDonald, Slav Petrov, Sampo Pyysalo,
  Natalia Silveira, Reut Tsarfaty, and Daniel Zeman. 2016.
\newblock Universal dependencies v1: A multilingual treebank collection.
\newblock In \emph{Proceedings of LREC}.

\bibitem[{Palmer et~al.(2005)Palmer, Gildea, and Kingsbury}]{Palmer:05}
Martha Palmer, Daniel Gildea, and Paul Kingsbury. 2005.
\newblock The {Proposition Bank}: An annotated corpus of semantic roles.
\newblock \emph{Computational Linguistics}, 31(1):71--106.

\bibitem[{Pennington et~al.(2014)Pennington, Socher, and
  Manning}]{pennington2014glove}
Jeffrey Pennington, Richard Socher, and Christopher~D. Manning. 2014.
\newblock {GloVe}: Global vectors for word representation.
\newblock In \emph{Proceedings of EMNLP}.

\bibitem[{Richardson et~al.(2018)Richardson, Berant, and
  Kuhn}]{Richardson2018-ov-naacl}
Kyle Richardson, Jonathan Berant, and Jonas Kuhn. 2018.
\newblock Polyglot semantic parsing in {APIs}.
\newblock In \emph{Proceedings of {NAACL}}.

\bibitem[{Roth and Lapata(2016)}]{Roth2016-fn}
Michael Roth and Mirella Lapata. 2016.
\newblock Neural semantic role labeling with dependency path embeddings.
\newblock {arXiv:1605.07515}.

\bibitem[{Srivastava et~al.(2015)Srivastava, Greff, and
  Schmidhuber}]{srivastava2015training}
Rupesh~Kumar Srivastava, Klaus Greff, and J{\"u}rgen Schmidhuber. 2015.
\newblock Training very deep networks.
\newblock In \emph{NIPS}.

\bibitem[{Swayamdipta et~al.(2016)Swayamdipta, Ballesteros, Dyer, and
  Smith}]{Swayamdipta2016-qt}
Swabha Swayamdipta, Miguel Ballesteros, Chris Dyer, and Noah~A. Smith. 2016.
\newblock Greedy, joint syntactic-semantic parsing with stack {LSTMs}.
\newblock In \emph{Proceedings of CoNLL}.

\bibitem[{Taul{\'e} et~al.(2008)Taul{\'e}, Mart{\'i}, and
  Recasens}]{taule2008ancora}
Mariona Taul{\'e}, M.~Ant{\`o}nia Mart{\'i}, and Marta Recasens. 2008.
\newblock {AnCora}: Multilevel annotated corpora for {Catalan} and {Spanish}.
\newblock In \emph{Proceedings of LREC}.

\bibitem[{Zhao et~al.(2009)Zhao, Chen, Kazama, Uchimoto, and
  Torisawa}]{Zhao2009}
Hai Zhao, Wenliang Chen, Jun'ichi Kazama, Kiyotaka Uchimoto, and Kentaro
  Torisawa. 2009.
\newblock Multilingual dependency learning: Exploiting rich features for
  tagging syntactic and semantic dependencies.
\newblock In \emph{Proceedings of CoNLL}.

\end{thebibliography}
\bibliographystyle{acl_natbib}

\end{document}